\documentclass[letterpaper, 10 pt, conference]{ieeeconf}  

\IEEEoverridecommandlockouts                              

\usepackage{graphics} 
\usepackage{epsfig} 
\usepackage{times} 
\usepackage[T1]{fontenc}
\usepackage[ruled,vlined,linesnumbered]{algorithm2e}
\usepackage{multirow}

\usepackage{amsmath, graphicx}
\usepackage{balance}
\usepackage{amssymb,amsbsy}
\usepackage{breqn,bbm,xcolor}
\usepackage{multirow}
\usepackage[symbol,bottom]{footmisc}
\usepackage{soul}
\usepackage{url}
\usepackage{booktabs} 
\usepackage{tikz}
\usepackage[export]{adjustbox}
\usepackage{subcaption}
\usepackage{float}
\usepackage{tabularx}
\usepackage{amssymb}

\title{\LARGE \bf
LapSurgie: Humanoid Robots Performing Surgery via Teleoperated Handheld Laparoscopy 
}

\author{Zekai Liang$^{1}$, Xiao Liang$^{1}$, Soofiyan Atar$^{1}$, Sreyan Das, Zoe Chiu$^{2}$,  Peihan Zhang$^{1}$, \\
 Calvin Joyce$^{1}$, Florian Richter$^{1}$, Shanglei Liu$^{3}$, Michael C. Yip$^{1}$, \IEEEmembership{Senior Member, IEEE} 
\thanks{$^{1}$Department of Electrical and Computer Engineering, University of California San Diego, La Jolla, CA 92093 USA.\ {\tt\small \{z9liang, x5liang, satar, pez004, cajoyce, frichter, yip\}@ucsd.edu}}%
\thanks{$^{2}$School of Electrical and Computer Engineering, Cornell University, Ithaca, NY 14853 USA.\ {\tt\small zc499@cornell.edu}}%
\thanks{$^{3}$UC San Diego Health, La Jolla, CA 92037 USA.\ {\tt\small s5liu@health.ucsd.edu}}}

\begin{document}

\maketitle
\thispagestyle{empty}
\pagestyle{empty}

\begin{abstract}
Robotic laparoscopic surgery has gained increasing attention in recent years for its potential to deliver more efficient and precise minimally invasive procedures. However, adoption of surgical robotic platforms remains largely confined to high-resource medical centers, exacerbating healthcare disparities in rural and low-resource regions. To close this gap, a range of solutions has been explored, from remote mentorship to fully remote telesurgery. Yet, the practical deployment of surgical robotic systems to underserved communities remains an unsolved challenge.
Humanoid systems offer a promising path toward deployability, as they can directly operate in environments designed for humans without extensive infrastructure modifications -- including operating rooms. In this work, we introduce LapSurgie, the first humanoid-robot-based laparoscopic teleoperation framework. The system leverages an inverse-mapping strategy for manual-wristed laparoscopic instruments that abides to remote center-of-motion constraints, enabling precise hand-to-tool control of off-the-shelf surgical laparoscopic tools without additional setup requirements. A control console equipped with a stereo vision system provides real-time visual feedback. Finally, a comprehensive user study across platforms demonstrates the effectiveness of the proposed framework and provides initial evidence for the feasibility of deploying humanoid robots in laparoscopic procedures.
\end{abstract}

\section{Introduction}

Minimally invasive surgery (MIS), particularly laparoscopic surgery, has revolutionized modern surgical practice by significantly reducing patient trauma, postoperative pain, and recovery time compared to traditional open procedures \cite{patil2024comparative}. Based on this, robot-assisted laparoscopic surgeries have drawn growing attention from researchers and surgeons for their potential to help reduce long-term operating fatigue \cite{shugaba2022should}, improve visual feedback \cite{wong2024manipulation}, provide extra dexterity and precision of motion \cite{wu2025efficacy}, and enable remote control capabilities \cite{biswas2023recent}.
However, surgical robot platforms have significant limitations as well, from million-dollar price tags \cite{mcbride2021detailed, burke2024robotic}, to large footprints \cite{kanji2021room, zhao2023robotic}, significant training requirements \cite{neri2024novel}, complex physical setups \cite{liu2024robotic}, and requirements for specialized rooms and management of surrounding equipment to accommodate the robot \cite{clanahan2023does}. 
All of these factors lead to the high capital, special training for personnel, deep facility resources, and time costs to a hospital.  Ultimately, few surgeons specialized in robotic surgery are accessible outside of large metropolitan hospitals, leaving patients at rural and low-resource hospitals and surgical centers without access to robotic surgical care \cite{WHO2025VirtualCareTelesurgery}.

\begin{figure}[t]
    \centerline{\includegraphics[width=0.85\linewidth,clip=true,trim={0mm 0mm 0mm 0mm}]{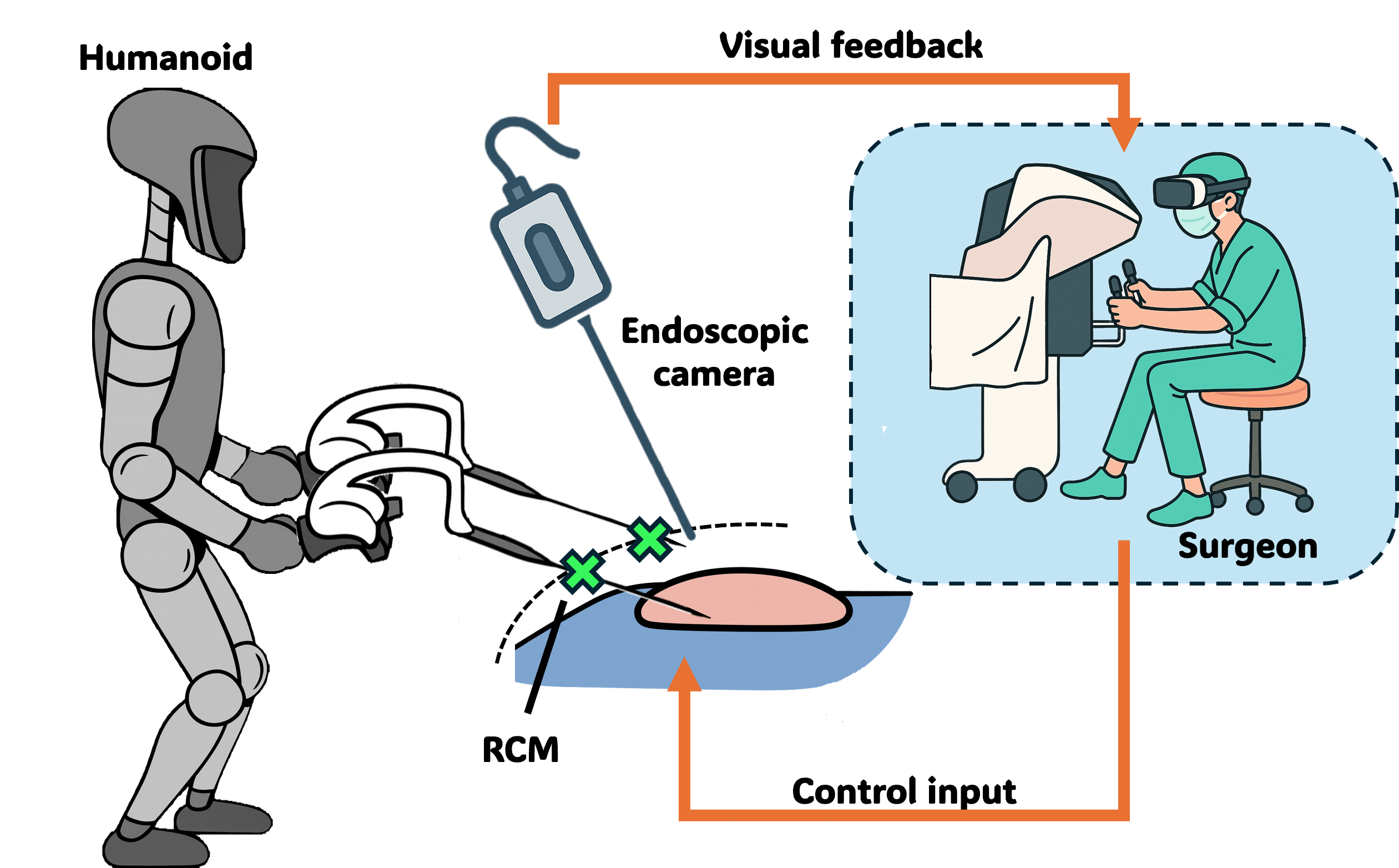}}
    \caption{Surgeons teleoperate humanoid robots to conduct surgical tasks. The proposed framework integrates direct control of manual laparoscopic instruments with remote center of motion (RCM) constraints and stereo vision feedback, enabling flexible deployment across diverse surgical settings.} 
    
    \label{demonstration}
    \vspace{-0.16in}
\end{figure}

\begin{figure*}[t]
    \centering
    \includegraphics[width=0.80\linewidth,clip=true,trim={0mm 0mm 0mm 0mm }]{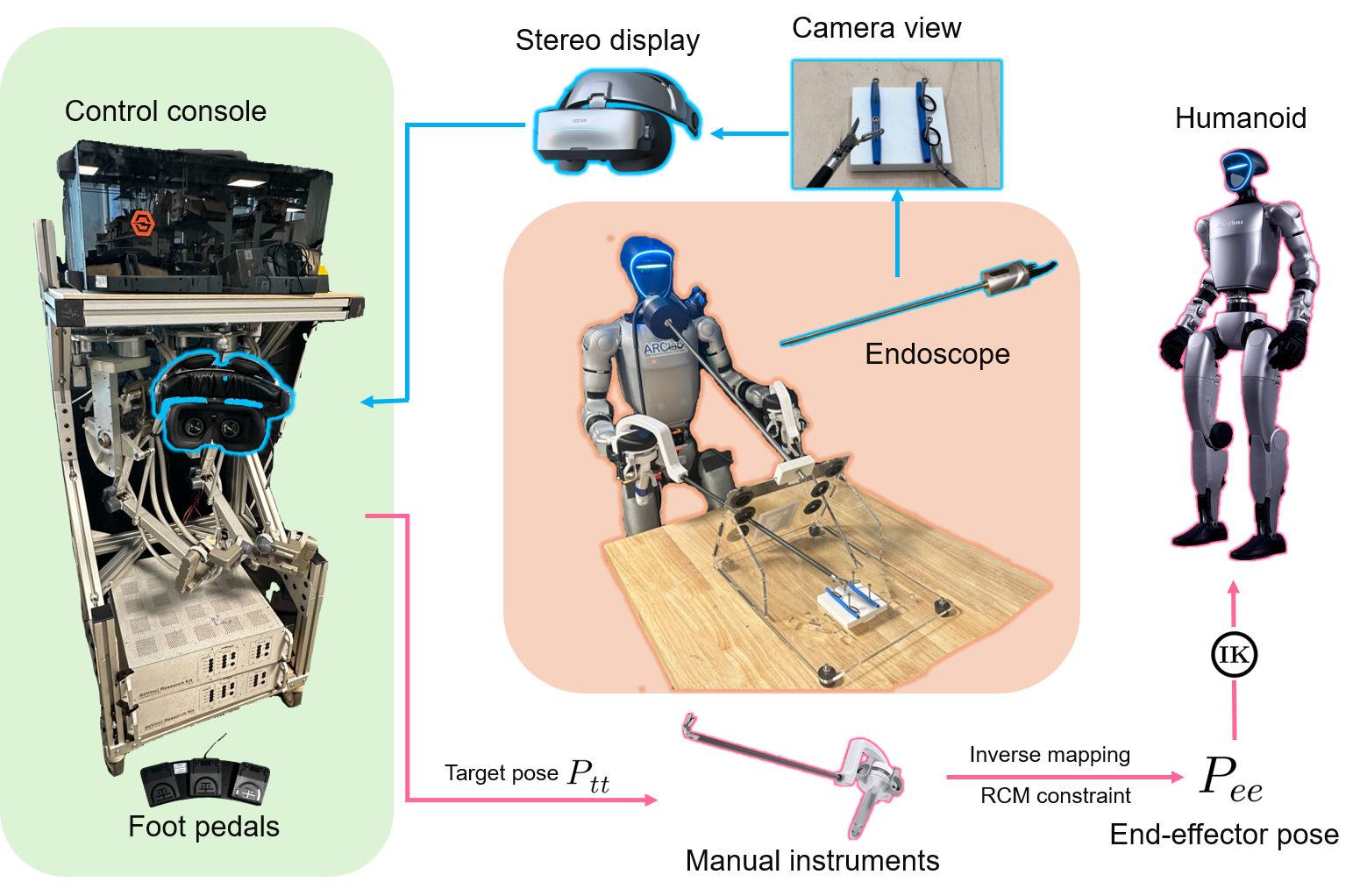}
    \caption{The overview of the humanoid-based laparoscopic framework. The target tool pose $\textbf{P}_{tt}$ is mapped from the control console handles and used as input for inverse-mapping control of manual laparoscopic instruments. A Cornerstone endoscope \cite{cornerstone_robotics} is mounted to provide real-time visual feedback, which is rendered through a stereo display to the operator during the procedure.}
    \label{fig:system}
    \vspace{-0.18in}
\end{figure*}

There have been efforts to utilize telementoring strategies to provide intraoperative guidance from remote experts \cite{shin2015novel, faris2023surgical}. More ambitious attempts have explored fully remote telesurgery \cite{anvari2005establishment, nguan2008robotic, barba2022remote}, accompanied by engineering solutions to mitigate the negative effects of latency, such as motion scaling \cite{richter2019motion, orosco2021compensatory, richter2021bench} and predictive displays \cite{shenai2011virtual, richter2019augmented}.
While these approaches directly address the challenge of providing specialized surgical expertise in low-resource settings, they overlook a more fundamental issue: the practical deployment and integration of surgical robotic systems themselves.

Humanoids, unlike surgical robots that are hyper-specialized for a single application, offer broad versatility due to their human-like form and compatibility with existing tools and environments. Their anthropomorphic design allows them to operate in hospitals without requiring dedicated operating suites, extensive infrastructure changes, or specialized equipment management \cite{atar2025humanoids}. In this way, humanoids have the future potential to lower deployment barriers and expand access to surgical robotic care, particularly in resource-limited or decentralized settings where conventional systems are impractical. At the same time, as general-purpose platforms, humanoids face the classic “jack-of-all-trades but master of none” dilemma, making it essential to identify domains where their adaptability provides unique value over bespoke systems. Recent work has explored humanoid teleoperation in a variety of domains \cite{he2024learning, penco2024mixed, li2025clone, myers2025child}, yet their role in surgery remains unexplored. To date, humanoids have never been deployed in real clinical operations.


To explore the potential of humanoid-based laparoscopic surgery, we introduce \textbf{LapSurgie} -- the first framework enabling humanoid robots to perform laparoscopic procedures.
The goal of this work is to investigate whether a general-purpose humanoid can support robot-assisted MIS without relying on specialized systems such as the da Vinci Surgical System.
Our teleoperation setup uses a G1 humanoid robot grasping commercially available laparoscopic instruments, controlled through the Master Tool Manipulators (MTMs) of the da Vinci Research Kit (dVRK) \cite{kazanzides2014dvrk}. These commercially available instruments include wrist degrees of freedom that are atypical and very difficult to use without extensive training, but that essentially reproduce the dexterity of robotic tools. By leveraging wristed laparoscopic tools and stereo vision, LapSurgie resolves a nontrivial mapping from human hand and finger motions to scaled, remapped, and wristed instrument control, thereby replicating the key benefits of conventional MIS robots.

To assess feasibility, we conduct a comparative user study across three conditions—manual laparoscopy, the dVRK surgical platform, and the humanoid system. The results provide initial evidence that humanoid robots can safely and effectively serve as a platform for laparoscopic surgery, opening a pathway toward deployable and flexible robotic surgical assistance.


    

\section{Related Works}
\subsection{Robotics-Assisted Surgery}
Robotic assistance has significantly advanced MIS, improving precision, reducing surgeon fatigue, and accelerating recovery times \cite{reddy2023advancements}. Surveys detail evolving trends in robotic MIS, including modular systems and emerging platforms in development \cite{ sarin2024upcoming}. The da Vinci Research Kit (dVRK) has catalyzed academic innovation in algorithmic control, skill benchmarking, and autonomy research \cite{kazanzides2014dvrk}. Noteworthy progress includes improved surgical scene understanding \cite{saha2025based, liang2025differentiable} and autonomous subtask execution in surgery\cite{saeidi2022autonomous, kim2025srt}. Investigations in emergency robotic surgery applications and enhanced dexterity via wristed instruments further underscore MIS innovation \cite{lunardi2024robotic, tu2025review}. Despite these advances, high costs, space constraints, and training requirements hinder widespread access, especially in resource-limited settings.

\subsection{Humanoid Telemanipulation}
Humanoids—designed for broad use—can adapt to human-centric environments such as hospitals without extensive infrastructure changes \cite{atar2025humanoids}. Recent breakthroughs in teleoperation have improved the fidelity and accessibility of humanoid control. For example,  \cite{he2024learning} achieves real-time whole-body control using only an RGB camera, while \cite{he2024omnih2o} allows multiple teleoperation modes and autonomy via GPT-4 integration. \cite{clone2025} introduces closed-loop, mixture-of-experts control to enable long-horizon, drift-resistant humanoid teleoperation. Other work is exploring haptic teleoperation for strength-intensive tasks \cite{purushottam2025heavy} and contact manipulation via deep tactile perception \cite{tact2025}. Although humanoid use in clinical operations (e.g., auscultation, palpation) has been tentatively explored \cite{atar2025humanoids}, humanoid-assisted laparoscopic surgery remains uncharted, marking a compelling opportunity for investigation\cite{yip2025robot}.

\section{Methodology}

The proposed framework overview is illustrated in Fig.\ref{fig:system}. At its core, the control console is integrated with an inverse-mapping control strategy for manual wristed laparoscopic instruments, enabling precise hand-to-tool control mapping. A remote center of motion constraint on top of this inverse mapping is applied to ensure keyhole positioning is unaffected by robot manipulation of the instruments. By coupling this natural manipulation paradigm with stereo vision feedback, the framework fully exploits the dexterity of humanoids while maintaining the intuitive ergonomics of conventional surgical robotic teleoperation.
In the following sections, \ref{sec:system} covers the development of the whole system, and \ref{sec:mapping} details the derivation of general laparoscopic tools' inverse mapping control with Remote-Center-of-Motion (RCM) constraint. 

\begin{figure}[t]
    \centering
    \includegraphics[width=0.5\linewidth]{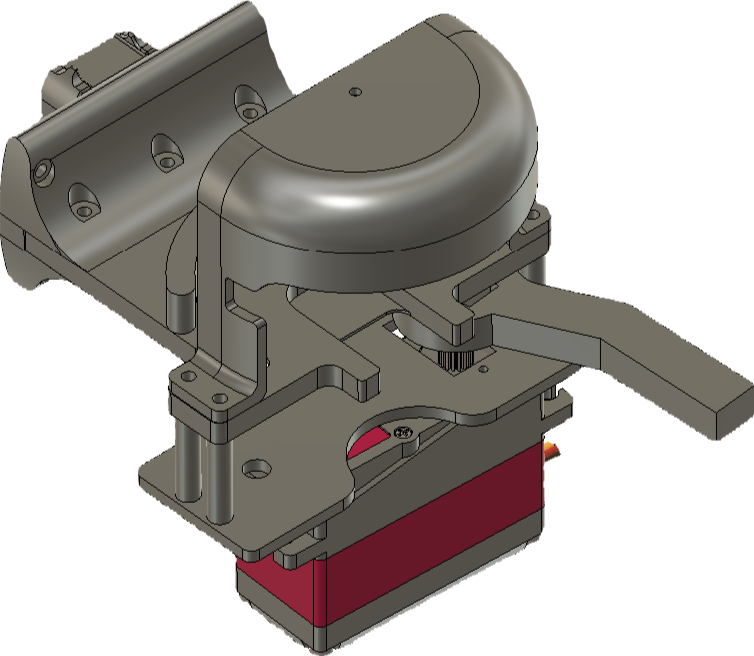}
    \caption{A coupling mount for a non-robotic laparoscopic instruments was created. This enabled the humanoid to operate the suite of wristed instruments directly, without modifications, enhancing the system’s generalizability across different laparoscopic needs. In this paper we used two types graspers developed by ArtiSential, held by the same mount.}
    \label{grasper model}
    \vspace{-0.14in}
\end{figure}

\subsection{Humanoid Laparoscopic Teleoperation System}
\label{sec:system}

As illustrated in Fig.\ref{fig:system}, this framework mounts two MTM modules \cite{kazanzides2014dvrk} on a lightweight mobile platform to serve as the system’s control interface. The proposed design integrates all essential control components, including the control workstation, MTM modules, foot pedals, and stereo vision unit into a single compact platform. The vision module incorporates a dual-1920p endoscopic camera and a GOOVIS G3 Max head-mounted stereo display, enabling immersive visual feedback for laparoscopic procedures. 

\begin{figure}[t]
    \centering
    \includegraphics[width=0.95\linewidth]{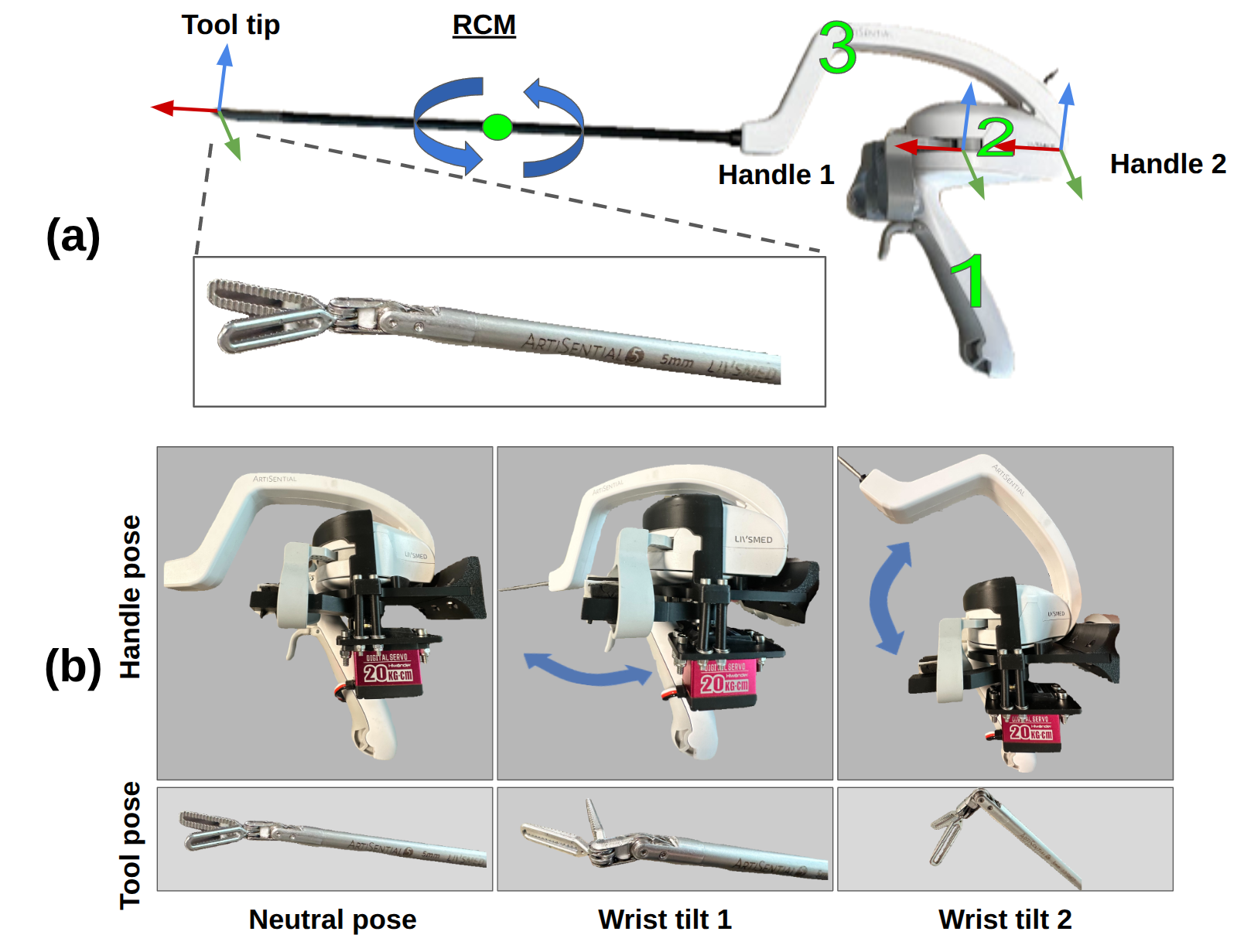}
    \caption{\textbf{(a)} The non-actuated laparoscopic tool consists of three main links from 1 to 3. We use three coordinate frames and RCM position in space to define the geometric connections of these components. The handle 1 and handle 2 frames are attached to link 1 and 3, respectively, and both are placed on the joints. \textbf{(b)} Since conventional laparoscopic tools are designed for human operation, the tool wrist motion is fully controlled by the relative angles between each part.}
    \label{FK}
    \vspace{-0.14in}
\end{figure}

On the humanoid side, the endoscope is mounted in front of the humanoid and streamed to the remote user to provide realistic laparoscopic views and real-time visual feedback. To enable the manipulation of general, non-robotic laparoscopic tools, we design a custom mount compatible with the wristed laparoscopic instruments, enabling grasping and full articulation of the wristed instruments  without modifying the original tools, as shown in Fig.\ref{grasper model}. The mount includes an interface for secure attachment to the humanoid hand, ensuring stability during manipulation, as shown in Fig.\ref{FK}. Identical mounts are installed on both hands of the humanoid.

The forceps are actuated via a servo-driven fingers that fit in the finger holes which open and close the laparoscopic instrument jaws. It maps the MTM input range to 0–60° (with 60° denoting a fully open state) and streams angle commands in real time to achieve precise grasper openness control.

Both control console and humanoid with instrument control is conducted via a ROS2 networked interface.

\subsection{Inverse Mapping Control of Manual Laparoscopic Tools}
\label{sec:mapping}


In this work, ArtiSential’s bipolar fenestrated forceps is used as the commercially available wristed laparoscopic instrument. ArtiSential is FDA 510(k)-cleared and commercialized by LivsMed; note that other wristed instruments are also available. Part of the kinematic chain, as described below, will involve understanding each instrument's kinematics, while establishing the RCM constraint will be instrument-agnostic.

Integrating manually-wristed laparoscopic instruments into robotic teleoperation typically introduces a much more complex passive kinematic chain, since these instruments are originally designed for direct human manipulation and lack actively actuated joints. As a result, the entire kinematic structure must be modeled based on the relative poses of its components. We define all coordinate frames of the instrument with respect to the robot base, as shown in Fig.\ref{FK}. The robot end-effector, grasper, and "handle 1" frames are treated as a rigid body without relative motion. 

\begin{figure}[t]
    \centering
    \includegraphics[width=0.95\linewidth]{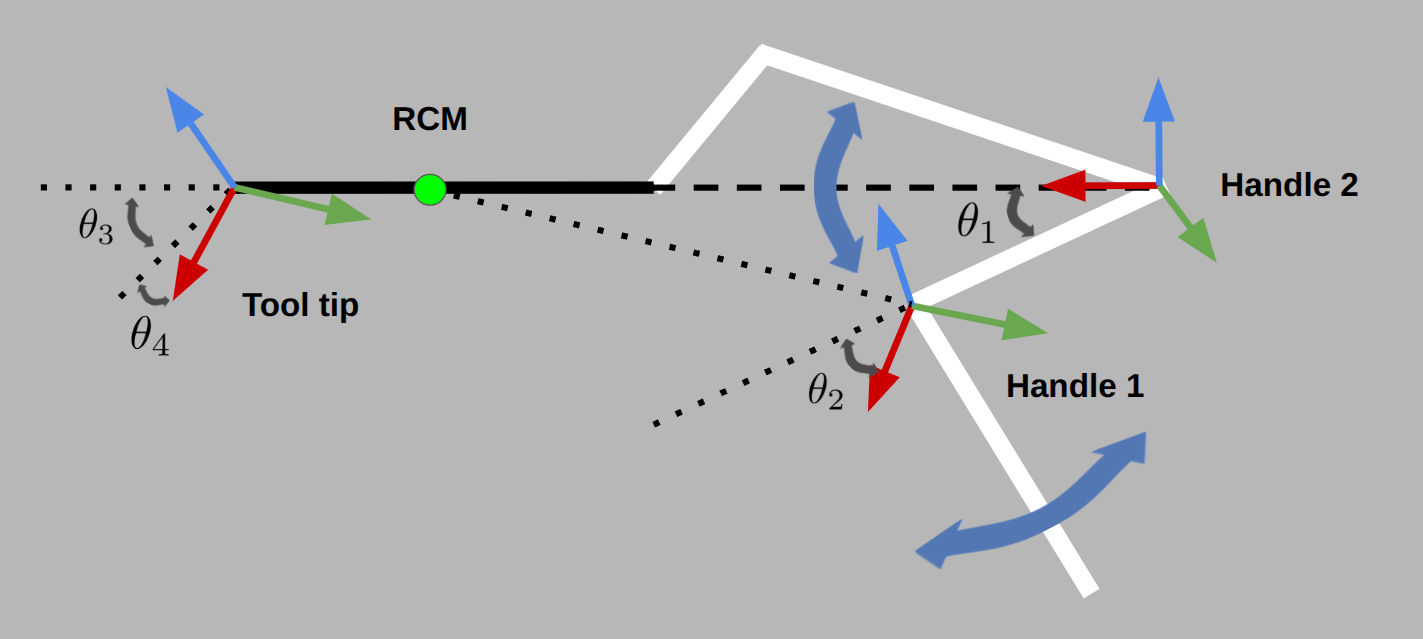}
    \caption{The relative orientation of the non-actuated instrument is modeled using two independent orthogonal rotations, $\theta_1$ and $\theta_2$, governed by both the handle pose and the RCM position in space.}
    \label{model}
    \vspace{-0.2in}
\end{figure}

To address this complexity, the passive geometry of the instruments is modeled as illustrated in Fig.\ref{model}. The variables $\theta_1$ and $\theta_2$ denote the relative handle angles with respect to the neutral position, while $\theta_3$ and $\theta_4$ are the relative angles of the tool tip. Using forward kinematics of the humanoid robot, the robot end-effector (wrist) pose is obtained as

\begin{equation}
    \textbf{P}_{ee} = f_{FK}(\boldsymbol{q}) \in \mathbb{R}^{4 \times 4},
\end{equation}
where $\boldsymbol{q}$ is the robot joint configuration. The handle 1 pose is obtained as

\begin{equation}
    \textbf{P}_{h1} = \textbf{P}_{ee}{}^{ee}\textbf{T}_{h1},
\end{equation}
with ${}^{ee}\textbf{T}_{h1} \in \mathbb{R}^{4 \times 4}$ being the fixed transform from the end-effector to handle 1.

The RCM position $\textbf{x}_{rcm} \in \mathbb{R}^3$ is calibrated with ArUco markers attached on the laparoscopy trainer board. Let $l_{0}$ denote the distance from handle~2 to the RCM, and $l_{12}$ the offset between handles 1 and 2, both physically measured directly from the tool. By the law of cosines, $\theta_1$ is computed as

\begin{equation}
    \theta_1 = \arccos\left(\frac{l_{12}^2 + l_0^2 - \|\textbf{x}_{rcm} - \textbf{x}_{h1}\|^2}{2l_{12}l_0}\right).
\end{equation}
The passive mechanism is further constrained by
\begin{equation}
    \|\textbf{x}_{h1} - \textbf{x}_{h2}\| = l_{12}, \quad
    \|\textbf{x}_{rcm} - \textbf{x}_{h2}\| = l_{0},
\end{equation}
and the perpendicularity condition
\begin{equation}
    (\textbf{R}_{h1}[0,0,1]^T)\cdot(\textbf{x}_{h1}-\textbf{x}_{h2})^T=0.
\end{equation}
Solving these yields the handle~2 position:
\begin{equation}
    \textbf{x}_{h2} = \textbf{x}_{h1} + l_{12} \cdot \frac{\textbf{n}_1 \times (\textbf{n}_1 \times \textbf{n}_2)}{\|\textbf{n}_1 \times (\textbf{n}_1 \times \textbf{n}_2)\|},
\end{equation}
where
\begin{equation}
    \textbf{n}_1 = \textbf{R}_{h1}[0,0,1]^T, \quad \textbf{n}_2 = \textbf{x}_{rcm} - \textbf{x}_{h1}.
\end{equation}
Extending from handle~2 to the tool tip, the remaining chain is treated as a rigid link with a known length $l_1$:
\begin{equation}
    \textbf{x}_{tt} = \textbf{x}_{h2} + \frac{l_1}{l_0}(\textbf{x}_{rcm}-\textbf{x}_{h2}).
\end{equation}
With $\textbf{x}_{h2}$ obtained, $\theta_2$ is given by
\begin{equation}
    \theta_2 = \arccos\left(\frac{\textbf{v}_1 \cdot \textbf{v}_2}{\|\textbf{v}_1\| \cdot \|\textbf{v}_2\|}\right),
\end{equation}
where
\begin{equation}
    \textbf{v}_1 = \textbf{x}_{h1}-\textbf{x}_{h2}, \quad 
    \textbf{v}_2 = \textbf{R}_{h1}[1,0,0]^T \; .
\end{equation}
Finally, the tool tip orientation is recovered as
\begin{equation}
    \textbf{R}_{tt} = \textbf{R}_{h1}\,\textbf{R}(-\theta'_2)\,\textbf{R}(-\theta'_1)\,\textbf{R}(\theta'_3)\,\textbf{R}(\theta'_4),
\end{equation}

\begin{equation}
    \theta_3 = k \theta_1, \quad
    \theta_4 = k \theta_2,
\end{equation}
where the scaling factor $k=2$ is found to be a gearing ratio within the tool's physical mechanism. The signed values $\theta'_1$ and $\theta'_2$ are defined by
\begin{equation}
    \theta'_1 = \frac{(z_{h1}-z_{rcm})}{\|z_{h1}-z_{rcm}\|}\cdot \theta_1, \quad
    \theta'_2 = \frac{(\textbf{v}_1 \times \textbf{v}_2)\cdot \textbf{n}_2}{\|(\textbf{v}_1 \times \textbf{v}_2)\cdot \textbf{n}_2\|}\cdot \theta_2
\end{equation}
Finally, the tool tip pose can be expressed as 
\begin{equation}
    \mathbf{P}_{tt} =
    \begin{bmatrix}
    \mathbf{R}_{tt} & \mathbf{t}_{tt}\\
    \mathbf{0}^\top & 1
    \end{bmatrix}.
\end{equation}

This formulation models the extended kinematic chain of manual wristed laparoscopic tools, allowing accurate recovery of both the position and orientation of the tool tip.
Given a desired tool tip pose $\mathbf{P}_{tt}' \in SE(3)$ and RCM location $\mathbf{x}_{\text{rcm}}$, the handle 1 pose $\mathbf{T}_{h1} \in SE(3)$ can be solved by minimizing a weighted residual with soft joint angle limits. A residual vector incorporating pose reprojection error and passive angle limits is defined as
\begin{equation}
\mathbf{r}(\mathbf{P}_{h1}) =
\begin{bmatrix}
w_t \big( \mathbf{x}_{tt} - \mathbf{x}_{tt}' \big) \\[4pt]
\mathrm{log}\!\left(\mathbf{R}_{tt}^{\top} \mathbf{R}_{tt}'\right) \\[4pt]
w_a \cdot \max(0,|\theta_1|-\theta_{\max}) \\[2pt]
w_a \cdot \max(0,|\theta_2|-\theta_{\max}\big)
\end{bmatrix},
\label{eq:ik_residual}
\end{equation}
where $\mathbf{x}_{tt}$ and $\mathbf{R}_{tt}$ denote the tool tip position and orientation obtained from the current handle 1 pose $\mathbf{P}_{h1}$. $w_t$ and $w_a$ are the scale factors of each term. The angle limit $\theta_{\max}$ is set as 45 degrees as a mechanical constraint. The inverse mapped handle 1 pose is solved by the Trust-Region Reflective (TRF) algorithm with a linear loss function:
\begin{equation}
\mathbf{P}_{h1}^\star = \arg\min_{\mathbf{P}_{h1}} \;\; \mathcal{L}\!\big(\mathbf{r}(\mathbf{P}_{h1})\big) \; .
\end{equation}
Thus, the target robot end-effector pose $\mathbf{P}_{ee}^\star$ can be obtained by adding a physical offset from handle 1 frame to the robot wrist frame $\textbf{t}_{offset}$:
\begin{equation}
    \mathbf{P}_{ee}^\star = \mathbf{P}_{h1}^\star \begin{bmatrix}
    \mathbf{I} & \mathbf{t}_{offset}\\
    \mathbf{0}^\top & 1
    \end{bmatrix} \; .
\end{equation}

\begin{figure*}[ht]
    \centering  
    \includegraphics[width=0.95\linewidth]{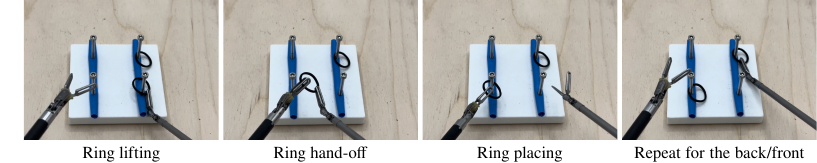}
    \caption{The user study task for each trial. It involves diverse tool motions and bi-manual operation, which presents high demand of the manipulation accuracy.}
    \label{task}
    \vspace{-0.1in}
\end{figure*}

\section{Experiments}
To thoroughly evaluate the performance of the proposed system and investigate the potential of this paradigm, we conduct a user study on the same task across three different platforms: Humanoid, da Vinci Research Kit (dVRK), and manual operation.  For the humanoid and manual experiments, the tasks are conducted within a laparoscopic trainer box that enforces remote center of motion (RCM) constraints (shown in Fig.\ref{fig:system}). 

\subsection{User Study Task}
\begin{figure}[t]
    \centering  
    \includegraphics[width=0.80\linewidth]{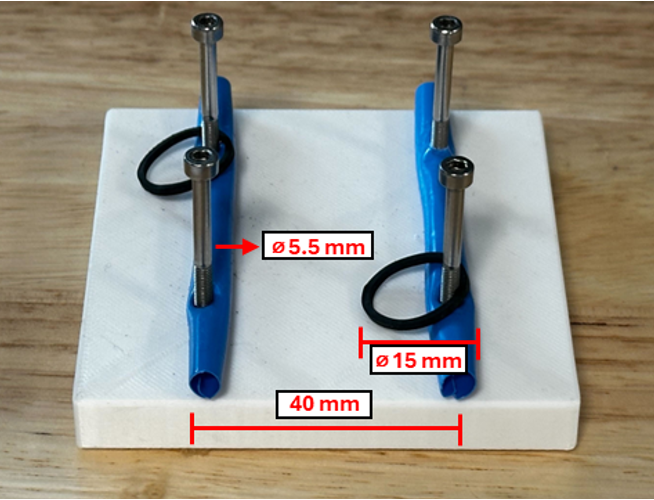}
    \caption{Experiment condition for the peg-transfer task. Straws are placed below o-rings to provide grasping clearance.}
    \label{fig:condition}
    \vspace{-0.14in}
\end{figure}

Following a similar pattern of \cite{richter2019motion}, the task for each trial is a standardized peg-transfer procedure. As illustrated in Fig.~\ref{fig:condition}, the experimental setup consists of a training board with four pegs arranged in a square configuration, each separated by 40\,mm. Two rubber o-rings are placed on the pegs as the task objects, with straws positioned beneath them to provide sufficient grasping clearance from the board surface. Each trial includes the following steps, as shown in Fig.\ref{task}. 

\begin{enumerate}
    \item Lift a rubber o-ring from either the front (back) left or right peg using the corresponding arm.
    \item Pass the o-ring to the other arm.
    \item Place the o-ring on the opposite front (back) peg.
    \item Repeat the same procedure for the back (front) pair of pegs.
\end{enumerate}


\begin{figure}[ht]
    \centering  
    \includegraphics[width=0.85\linewidth]{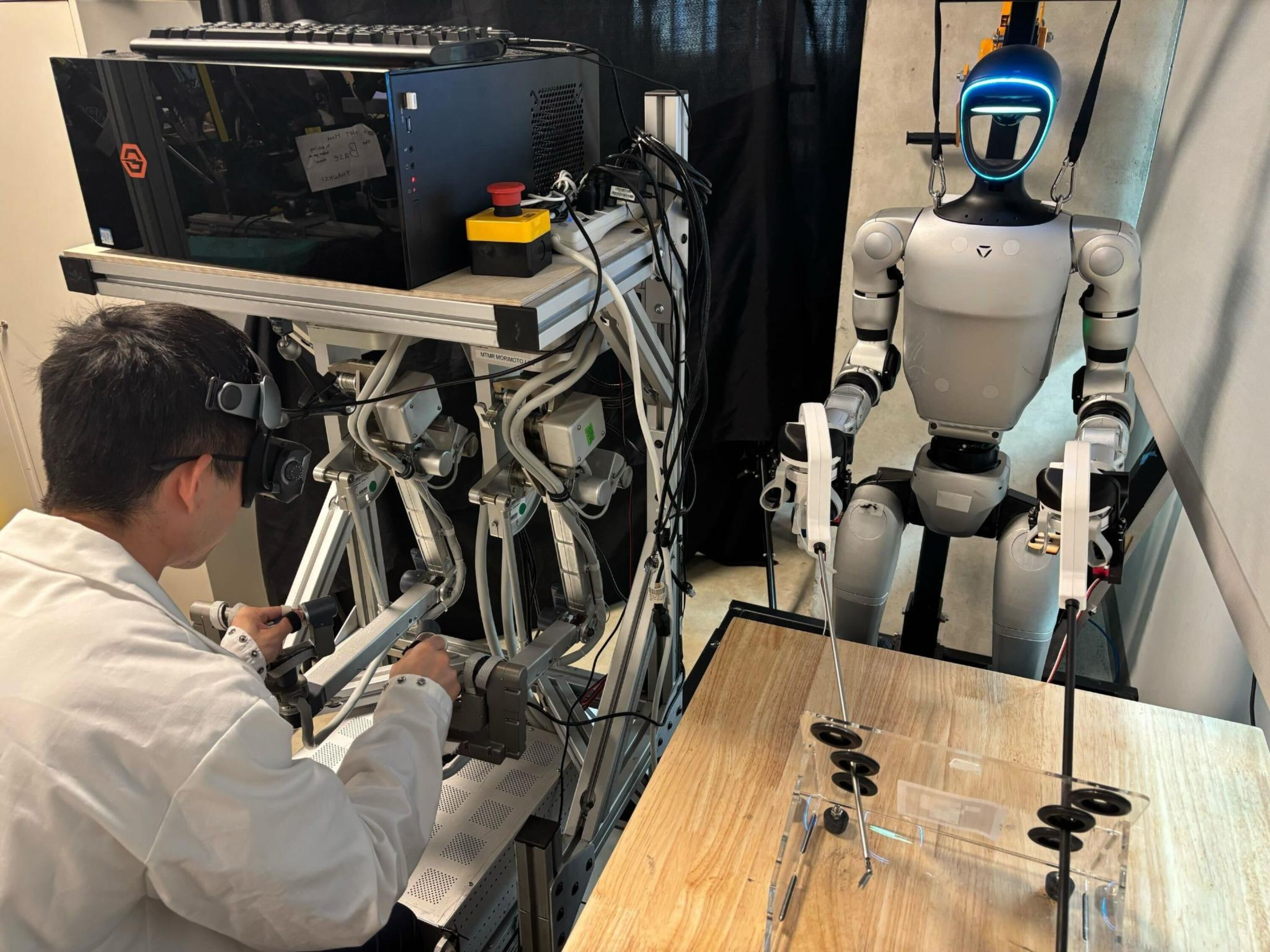}
    \caption{Participants performing laparoscopic operations with the developed humanoid-based surgical framework.}
    \label{fig:operation}
    \vspace{-0.14in}
\end{figure}

\begin{table}[ht]
\centering
\caption{Weighted error scores for different error types.}
\label{tab:error_weights}
\begin{tabular}{|l|c|}
\hline
\textbf{Error Type} & \textbf{Weight} \\ \hline
Failed pick up                        & 2  \\ \hline
Stretch ring on pegs                     & 2  \\ \hline
Stretch ring during hand-off             & 4  \\ \hline
Drop ring (outside of pegs)      & 5  \\ \hline
Collision (pegs \& ground \& tools)     & 3 \\ \hline
Straw displacement               & 3 \\ \hline
\end{tabular}
\vspace{-0.17in}
\end{table}

\begin{table*}[t]
\centering
\caption{Per-trial weighted error and time statistics across novices and surgeons. $p$-values are from two-sided paired $t$-tests. Humanoid platform achieves comparable operation accuracy with dVRK, while with longer completion times.}
\label{tab:combined_stats}
\begin{tabular}{l l@{\hskip 0.8cm}c @{\hskip 0.8cm}c @{\hskip 0.8cm}c@{\hskip 0.8cm}c @{\hskip 0.8cm}c @{\hskip 0.8cm}c}
\toprule
& & \multicolumn{3}{c}{\textbf{Weighted Error}} & \multicolumn{3}{c}{\textbf{Time (s)}} \\
\cmidrule(lr){3-5} \cmidrule(lr){6-8}
\textbf{Group} & \textbf{Platform} & mean $\pm$ std & $p$ vs. Manual & $p$ vs. dVRK & mean $\pm$ std & $p$ vs. Manual & $p$ vs. dVRK \\
\midrule
 Novices  & Manual   & $7.91 \pm 5.64$ & --     & 0.096  & $75.43 \pm 32.66$ & --      & 0.006 \\
             & Humanoid & $5.49 \pm 3.35$ & 0.091  & 0.575  & $89.23 \pm 38.91$ & 0.220   & $8.46 \times 10^{-5}$ \\
             & dVRK     & $5.00 \pm 2.83$ & 0.096  & --     & $45.37 \pm 17.93$ & 0.006   & --  \\
\midrule
 Surgeons & Manual   & $4.25 \pm 2.47$ & --     & 0.597  & $44.80 \pm 9.30$  & --      & 0.022 \\
             & Humanoid & $2.44 \pm 2.03$ & 0.109  & 0.534  & $66.75 \pm 15.92$ & 0.434   & 0.368 \\
             & dVRK     & $3.56 \pm 3.80$ & 0.597  & --     & $39.16 \pm 9.58$  & 0.022   & -- \\
\bottomrule
\end{tabular}
\vspace{-0.10in}
\end{table*}

\begin{table*}[t]
\centering
\caption{Error type occurrences per platform across all participants.}
\label{tab:errors occurences}
\begin{tabular}{lcccccc}
\toprule
\textbf{Platform} & \textbf{Failed pick up} & \textbf{Stretch ring on pegs} & \textbf{Stretch ring during hand-off} & \textbf{Drop ring} & \textbf{Collision} & \textbf{Straw displacement} \\
\midrule
dVRK     & $3.65 \pm 2.14$ & $0.88 \pm 0.98$ & $0.33 \pm 0.94$ & $0.38 \pm 0.69$ & $8.19 \pm 5.20$ & $0.64 \pm 0.74$ \\
Humanoid & $6.27 \pm 5.22$ & $3.35 \pm 3.39$ & $0.48 \pm 0.65$ & $1.22 \pm 1.44$ & $3.70 \pm 4.03$ & $0.34 \pm 0.83$ \\
Manual   & $6.25 \pm 4.86$ & $6.16 \pm 5.87$ & $1.04 \pm 1.17$ & $0.74 \pm 1.07$ & $8.49 \pm 6.22$ & $0.31 \pm 0.51$ \\
\bottomrule
\end{tabular}
\vspace{-0.10in}
\end{table*}

\begin{figure*}[ht]
    
    \centering  
    \includegraphics[width=0.90\linewidth]{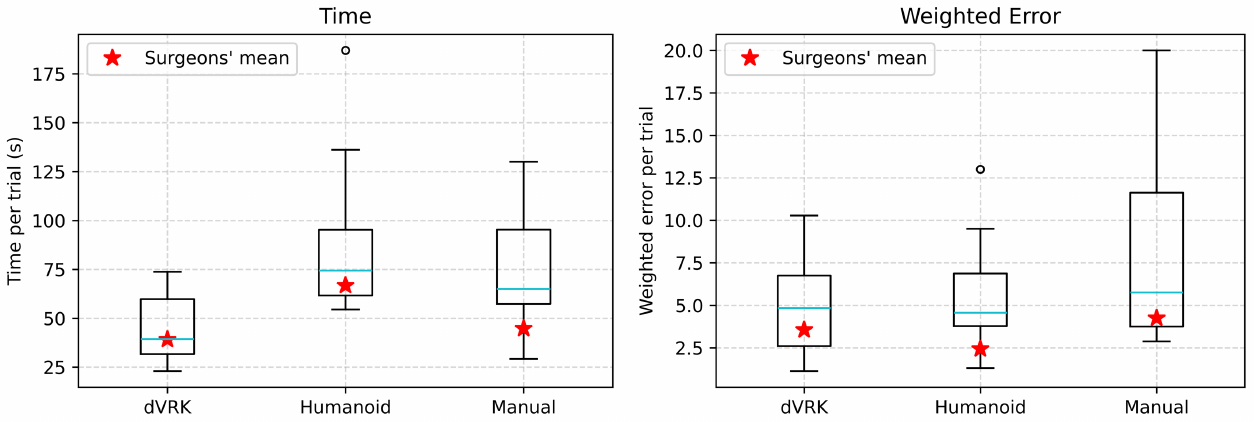}
    \caption{Per-trial weighted error and completion time for novices (boxplots) and the red stars indicate the mean performance of the surgeons.
    The observed error rates on the humanoid-based laparoscopic platform are on par with those observed from the established dVRK system.
    However, the completion time of the Humanoid platform lags behind dVRK.
    }
    \label{fig:error boxplot}
    \vspace{-0.16in}
\end{figure*}

\subsection{User study procedure}
Each participant would go through a structured procedure for the result recording:
\begin{enumerate}
     \item \textbf{Pre-experiment:} Prior to the official trials, all participants undergo a standardized training session to familiarize themselves with each platform. For the robotic systems, this includes approximately five minutes of practice to learn the fundamental operations, such as handle motion control, foot pedal usage, , the vision module, as well as safety and cautions. For the manual condition, participants spend an equivalent amount of time practicing the basic manipulation of wristed laparoscopic tools. After the initial familiarization, each participant completes the operation task twice before proceeding to the official trials.
    
    \item \textbf{Trials \& recording:} After the pre-experiment training, participants conduct the official trials while the recording starts. For each platform, the participants perform the same operation task 8 times in total as a good balance of obtaining informative performance data and minimizing the effect of fatigue. For each trial, participants' operation performance and time are recorded for evaluation. When recording each trial, the timer starts as the tools touch the first ring. An example of the operation is shown in Fig.~\ref{fig:operation}.

    \item \textbf{Platform rotation:} Participants repeat step 1-2 on all three platforms (Humanoid, dVRK, and manual) with randomized orders. The trials of different platforms were separated for each participants to minimize potential performance degradation due to long operation fatigue. 
    \item \textbf{Evaluation metrics:} In this study, there is no hard failure, instead the performance is evaluated by weighted error scores and total time cost. As shown in Table~\ref{tab:error_weights}, common operation error types are assigned corresponding weights, including failed ring pick-up,  ring over-stretching on pegs or during hand-off, ring drops, collisions during operation, and straw displacement.
    
    After completing all the required operations, participants would fill out a post-experiment questionnaire to evaluate their experience on three platforms. The TLX \cite{hart1988development} style questionnaire consists of the following terms with a scale 1-10:

    \begin{itemize}
    \item \textit{Mental Demand} – the level of mental effort required to complete the task.
    \item \textit{Physical Demand} – the amount of physical effort and fatigue experienced.
    \item \textit{Motion Accuracy} – the perceived precision and controllability of tool movements.
    \item \textit{Feedback Quality} – the clarity and usefulness of visual/control feedback during operation.
    \item \textit{Overall Performance} – the participant’s overall experience of the task execution.
\end{itemize}

\end{enumerate}

\begin{figure*}[ht]
    \centering  
    \includegraphics[width=0.75\linewidth,clip=true,trim={0mm 8mm 0mm 0mm }]{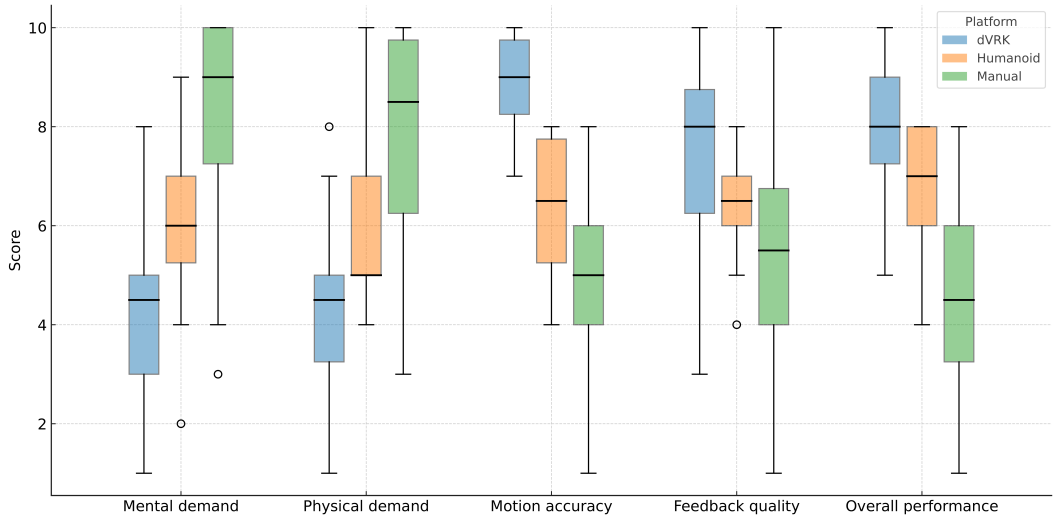}
    \vspace{-0.08in}
    \caption{Post-study questionnaire results for dVRK, Humanoid and Manual user study. The results demonstrate that the proposed humanoid-based laparoscopic platform requires significantly less mental and physical demand compared with manual operation, while achieving more precise manipulation and overall performance.
    Compared with dVRK, the Humanoid platform is rated slightly higher in mental and physical demand but still demonstrates strong performance in motion accuracy, feedback quality, and overall execution.}
    \label{post study}
    \vspace{-0.18in}
\end{figure*}

\section{Results and discussion}

A total of 14 participants took part in the user study, including 2 professional surgeons and 12 novices who had not received any surgical operation training previously.  

\subsection{Performance evaluation}
The statistics of both groups' performance across platforms are presented in the Table \ref{tab:combined_stats}, which includes the per-trial weighted error scores and operation times. The boxplot distribution is shown in Fig.\ref{fig:error boxplot}. Additionally, each error type's occurrences are reported in Table \ref{tab:errors occurences} for more straightforward comparison.

In the novices group, dVRK platform achieves the lowest average weighted error and the fastest execution time. Manual operation performance, while relatively fast ($78.14 \pm 50.52$\,s), exhibits the highest error rates  and inconsistency ($8.19 \pm 9.67$). The Humanoid platform yields comparable accuracy ($5.78 \pm 6.47$) to the dVRK, but novices required substantially more time to complete tasks ($p < 0.05$).

In the surgeons group, overall performance is more accurate and consistent across all platforms compared to the novices. Surgeons complete tasks most quickly on the dVRK, yet they achieve their lowest error rates on the Humanoid platform. Time-wise, the statistics show the similar trend as in the novices group: dVRK still yields the fastest execution while humanoid takes the relatively longest time. 

These results collectively demonstrate the feasibility of employing the Humanoid platform for laparoscopic surgical tasks. In the novices group, their error rates on the Humanoid are already on par with those observed on the dVRK and substantially lower than in manual operation. Moreover, the fact that expert surgeons achieve the lowest error rates on the Humanoid highlights its potential to support highly precise manipulation once sufficient expertise is established. Although efficiency on the Humanoid currently lags behind the dVRK, which is largely due to the additional control scheme for the general surgical tools, the observed accuracy advantages suggest that with further training and system refinement, the Humanoid platform can serve as a viable and promising alternative for complex surgical procedures. Manual operation, in contrast, achieves faster execution because of the absence of any system control loop latency.


\subsection{Post-study questionnaire}
The results of the post-participation questionnaire are shown in the Fig. \ref{post study}, including the statistics of users' feedback of the three platforms. The results demonstrate that the proposed humanoid-based laparoscopic platform requires significantly less mental and physical demand compared with manual operation, while achieving more precise manipulation and overall performance. Compared with dVRK, the Humanoid platform is rated slightly higher in mental and physical demand but still demonstrates strong performance in motion accuracy, feedback quality, and overall execution. From discussions with participants, particularly surgeons, it was noted that the humanoid platform substantially reduces cognitive load compared with manual operation through the intuitive hand-to-tool mapping control. When compared with the dVRK system, the main gap lies in higher control latency and reduced precision. We believe these challenges can be effectively resolved in the future through improved hardware design, more efficient control schemes, and more accurate geometric modeling.

\section{Conclusion}
In this work, we propose the first study on humanoid-based robotic surgery. We explore the feasibility of deploying humanoid + general laparoscopic instruments on surgical tasks, as a potential alternative to conventional surgical robot platforms with better deployability and accessibility. The experiment results demonstrate the proposed framework achieves operation accuracy comparable to the current gold standard dVRK, while significantly outperforming manual operation. Although the current system still trails the dVRK in terms of efficiency, this gap can be attributed to hardware design and control complexity. With continued refinement of tool integration and control strategy, humanoid-based surgical platforms hold strong potential to provide a flexible, scalable, and cost-effective alternative for advanced minimally invasive procedures. 






\balance
\bibliographystyle{ieeetr}
\bibliography{references}

@article{yip2025robot,
  title={The robot will see you now: Foundation models are the path forward for autonomous robotic surgery},
  author={Yip, Michael},
  journal={Science Robotics},
  volume={10},
  number={104},
  pages={eadt0684},
  year={2025},
  publisher={American Association for the Advancement of Science}
}

@article{shin2015novel,
  title={A novel interface for the telementoring of robotic surgery},
  author={Shin, Daniel H and Dalag, Leonard and Azhar, Raed A and Santomauro, Michael and Satkunasivam, Raj and Metcalfe, Charles and Dunn, Matthew and Berger, Andre and Djaladat, Hooman and Nguyen, Mike and others},
  journal={BJU international},
  volume={116},
  number={2},
  pages={302--308},
  year={2015},
  publisher={Wiley Online Library}
}

@article{faris2023surgical,
  title={Surgical tele-mentoring using a robotic platform: initial experience in a military institution},
  author={Faris, Hunter and Harfouche, Cyril and Bandle, Jesse and Wisbach, Gordon},
  journal={Surgical Endoscopy},
  volume={37},
  number={12},
  pages={9159--9166},
  year={2023},
  publisher={Springer}
}

@article{anvari2005establishment,
  title={Establishment of the world's first telerobotic remote surgical service: for provision of advanced laparoscopic surgery in a rural community},
  author={Anvari, Mehran and McKinley, Craig and Stein, Harvey},
  journal={Annals of surgery},
  volume={241},
  number={3},
  pages={460--464},
  year={2005},
  publisher={LWW}
}

@article{nguan2008robotic,
  title={Robotic pyeloplasty using internet protocol and satellite network-based telesurgery},
  author={Nguan, CY and Morady, R and Wang, C and Harrison, D and Browning, D and Rayman, R and Luke, PPW},
  journal={The International Journal of Medical Robotics and Computer Assisted Surgery},
  volume={4},
  number={1},
  pages={10--14},
  year={2008},
  publisher={Wiley Online Library}
}

@article{orosco2021compensatory,
  title={Compensatory motion scaling for time-delayed robotic surgery},
  author={Orosco, Ryan K and Lurie, Benjamin and Matsuzaki, Tokio and Funk, Emily K and Divi, Vasu and Holsinger, F Christopher and Hong, Steven and Richter, Florian and Das, Nikhil and Yip, Michael},
  journal={Surgical endoscopy},
  volume={35},
  number={6},
  pages={2613--2618},
  year={2021},
  publisher={Springer}
}

@article{barba2022remote,
  title={Remote telesurgery in humans: a systematic review},
  author={Barba, Patrick and Stramiello, Joshua and Funk, Emily K and Richter, Florian and Yip, Michael C and Orosco, Ryan K},
  journal={Surgical endoscopy},
  volume={36},
  number={5},
  pages={2771--2777},
  year={2022},
  publisher={Springer}
}

@inproceedings{richter2021bench,
  title={From bench to bedside: The first live robotic surgery on the dvrk to enable remote telesurgery with motion scaling},
  author={Richter, Florian and Funk, Emily K and Park, Won Seo and Orosco, Ryan K and Yip, Michael C},
  booktitle={2021 International Symposium on Medical Robotics (ISMR)},
  pages={1--7},
  year={2021},
  organization={IEEE}
}

@article{shenai2011virtual,
  title={Virtual interactive presence and augmented reality (VIPAR) for remote surgical assistance},
  author={Shenai, Mahesh B and Dillavou, Marcus and Shum, Corey and Ross, Douglas and Tubbs, Richard S and Shih, Alan and Guthrie, Barton L},
  journal={Operative Neurosurgery},
  volume={68},
  pages={ons200--ons207},
  year={2011},
  publisher={LWW}
}

@inproceedings{richter2019augmented,
  title={Augmented reality predictive displays to help mitigate the effects of delayed telesurgery},
  author={Richter, Florian and Zhang, Yifei and Zhi, Yuheng and Orosco, Ryan K and Yip, Michael C},
  booktitle={2019 international conference on robotics and automation (ICRA)},
  pages={444--450},
  year={2019},
  organization={IEEE}
}

@article{atar2025humanoids,
  title={Humanoids in Hospitals: A Technical Study of Humanoid Surrogates for Dexterous Medical Interventions},
  author={Atar, Soofiyan and Liang, Xiao and Joyce, Calvin and Richter, Florian and Ricardo, Wood and Goldberg, Charles and Suresh, Preetham and Yip, Michael},
  journal={arXiv preprint arXiv:2503.12725},
  year={2025}
}

@inproceedings{he2024learning,
  title={Learning human-to-humanoid real-time whole-body teleoperation},
  author={He, Tairan and Luo, Zhengyi and Xiao, Wenli and Zhang, Chong and Kitani, Kris and Liu, Changliu and Shi, Guanya},
  booktitle={2024 IEEE/RSJ International Conference on Intelligent Robots and Systems (IROS)},
  pages={8944--8951},
  year={2024},
  organization={IEEE}
}

@article{li2025clone,
  title={CLONE: Closed-Loop Whole-Body Humanoid Teleoperation for Long-Horizon Tasks},
  author={Li, Yixuan and Lin, Yutang and Cui, Jieming and Liu, Tengyu and Liang, Wei and Zhu, Yixin and Huang, Siyuan},
  journal={arXiv preprint arXiv:2506.08931},
  year={2025}
}

@article{myers2025child,
  title={CHILD (Controller for Humanoid Imitation and Live Demonstration): a Whole-Body Humanoid Teleoperation System},
  author={Myers, Noboru and Kwon, Obin and Yamsani, Sankalp and Kim, Joohyung},
  journal={arXiv preprint arXiv:2508.00162},
  year={2025}
}

@article{penco2024mixed,
  title={Mixed reality teleoperation assistance for direct control of humanoids},
  author={Penco, Luigi and Momose, Kazuhiko and McCrory, Stephen and Anderson, Dexton and Kitchel, Nicholas and Calvert, Duncan and Griffin, Robert J},
  journal={IEEE Robotics and Automation Letters},
  volume={9},
  number={2},
  pages={1937--1944},
  year={2024},
  publisher={IEEE}
}

@article{sarin2024upcoming,
  title={Upcoming multi-visceral robotic surgery systems: a SAGES review},
  author={Sarin, A.},
  journal={Surgical Endoscopy},
  year={2024},
}

@inproceedings{kazanzides2014dvrk,
  title={An Open-Source Research Kit for the da Vinci Surgical System (dVRK)},
  author={Kazanzides, Peter and others},
  booktitle={ICRA},
  year={2014},
}

@article{saeidi2022autonomous,
  title={Autonomous robotic laparoscopic surgery for intestinal anastomosis},
  author={Saeidi, Hamed and Opfermann, Justin D and Kam, Michael and Wei, Shuwen and L{\'e}onard, Simon and Hsieh, Michael H and Kang, Jin U and Krieger, Axel},
  journal={Science robotics},
  volume={7},
  number={62},
  pages={eabj2908},
  year={2022},
  publisher={American Association for the Advancement of Science}
}

@article{kim2025srt,
  title={SRT-H: A hierarchical framework for autonomous surgery via language-conditioned imitation learning},
  author={Kim, Ji Woong and Chen, Juo-Tung and Hansen, Pascal and Shi, Lucy Xiaoyang and Goldenberg, Antony and Schmidgall, Samuel and Scheikl, Paul Maria and Deguet, Anton and White, Brandon M and Tsai, De Ru and others},
  journal={Science robotics},
  volume={10},
  number={104},
  pages={eadt5254},
  year={2025},
  publisher={American Association for the Advancement of Science}
}

@article{lunardi2024robotic,
  title={Robotic technology in emergency general surgery cases in the era of minimally invasive surgery},
  author={Lunardi, Nicole and Abou-Zamzam, Aida and Florecki, Katherine L and Chidambaram, Swathikan and Shih, I-Fan and Kent, Alistair J and Joseph, Bellal and Byrne, James P and Sakran, Joseph V},
  journal={JAMA surgery},
  volume={159},
  number={5},
  pages={493--499},
  year={2024},
  publisher={American Medical Association}
}

@article{tu2025review,
  title={A review of wrist mechanism design and the application in gastrointestinal minimally invasive surgery of multi-degree-of-freedom surgical laparoscopic instruments},
  author={Tu, Yisi and Jiang, Jianhao and Huang, Jingyun and Sui, Jianbo and Yang, Shibin},
  journal={Surgical Endoscopy},
  volume={39},
  number={1},
  pages={99--121},
  year={2025},
  publisher={Springer}
}

@article{he2024omnih2o,
  title={Omnih2o: Universal and dexterous human-to-humanoid whole-body teleoperation and learning},
  author={He, Tairan and Luo, Zhengyi and He, Xialin and Xiao, Wenli and Zhang, Chong and Zhang, Weinan and Kitani, Kris and Liu, Changliu and Shi, Guanya},
  journal={arXiv preprint arXiv:2406.08858},
  year={2024}
}

@article{clone2025,
  title={CLONE: Closed-Loop Whole-Body Humanoid Teleoperation for Long-Horizon Tasks},
  author={Li, Yixuan and Lin, Yutang and Cui, Jieming and Liu, Tengyu and Liang, Wei and Zhu, Yixin and Huang, Siyuan},
  journal={arXiv preprint arXiv:2506.08931},
  year={2025}
}

@inproceedings{purushottam2025heavy,
  title={Heavy lifting tasks via haptic teleoperation of a wheeled humanoid},
  author={Purushottam, Amartya and Yan, Jack and Xu, Christopher and Ramos, Joao},
  booktitle={2025 IEEE-RAS 24th International Conference on Humanoid Robots (Humanoids)},
  pages={345--350},
  year={2025},
  organization={IEEE}
}

@article{tact2025,
  title={TACT: Humanoid Whole-body Contact Manipulation through Deep Imitation Learning with Tactile Modality},
  author={Murooka, Masaki and Hoshi, Takahiro and Fukumitsu, Kensuke and Masuda, Shimpei and Hamze, Marwan and Sasaki, Tomoya and Morisawa, Mitsuharu and Yoshida, Eiichi},
  journal={IEEE Robotics and Automation Letters},
  year={2025},
  publisher={IEEE}
}

@article{reddy2023advancements,
  title={Advancements in robotic surgery: a comprehensive overview of current utilizations and upcoming frontiers},
  author={Reddy, Kavyanjali and Gharde, Pankaj and Tayade, Harshal and Patil, Mihir and Reddy, Lucky Srivani and Surya, Dheeraj and srivani Reddy, Lucky},
  journal={Cureus},
  volume={15},
  number={12},
  year={2023},
  publisher={Cureus}
}

@inproceedings{richter2019motion,
  title={Motion scaling solutions for improved performance in high delay surgical teleoperation},
  author={Richter, Florian and Orosco, Ryan K and Yip, Michael C},
  booktitle={2019 International Conference on Robotics and Automation (ICRA)},
  pages={1590--1595},
  year={2019},
  organization={IEEE}
}

@article{shugaba2022should,
  title={Should all minimal access surgery be robot-assisted? A systematic review into the musculoskeletal and cognitive demands of laparoscopic and robot-assisted laparoscopic surgery},
  author={Shugaba, Abdul and Lambert, Joel E and Bampouras, Theodoros M and Nuttall, Helen E and Gaffney, Christopher J and Subar, Daren A},
  journal={Journal of Gastrointestinal Surgery},
  volume={26},
  number={7},
  pages={1520--1530},
  year={2022},
  publisher={Elsevier}
}

@article{wu2025efficacy,
  title={Efficacy and safety of robotic-assisted versus endoscopic-assisted axillary lymph node dissection in node-positive breast cancer: a retrospective comparative study},
  author={Wu, Zhijie and Liu, Qiwen and Li, Zongyan and Chen, Zuxiao and Wu, Yongxin and Luo, Yunxiang and Wei, Lina and Hu, Qiongyu and Li, Haiyan},
  journal={World Journal of Surgical Oncology},
  volume={23},
  number={1},
  pages={179},
  year={2025},
  publisher={Springer}
}

@article{wong2024manipulation,
  title={Manipulation ergonomics and robotic surgery—a narrative review},
  author={Wong, Shing Wai and Crowe, Philip},
  journal={Annals of Laparoscopic and Endoscopic Surgery},
  volume={9},
  year={2024},
  publisher={AME Publishing Company}
}

@article{biswas2023recent,
  title={Recent advances in robot-assisted surgical systems},
  author={Biswas, Pradipta and Sikander, Sakura and Kulkarni, Pankaj},
  journal={Biomedical Engineering Advances},
  volume={6},
  pages={100109},
  year={2023},
  publisher={Elsevier}
}

@article{patil2024comparative,
  title={Comparative analysis of laparoscopic versus open procedures in specific general surgical interventions},
  author={Patil Jr, Mihir and Gharde, Pankaj and Reddy, Kavyanjali and Nayak, Krushank and Patil, Mihir},
  journal={Cureus},
  volume={16},
  number={2},
  year={2024},
  publisher={Cureus}
}

@misc{WHO2025VirtualCareTelesurgery,
  author       = {{World Health Organization} and {Society of Robotic Surgery}},
  title        = {WHO and Society of Robotic Surgery launch health innovation initiative to expand access to virtual care and telesurgery},
  howpublished = {WHO News Release},
  month        = aug,
  day          = {8},
  year         = {2025},
  url          = {https://www.who.int/news/item/08-08-2025-who-and-society-of-robotic-surgery-launch-health-innovation-initiative-to-expand-access-to-virtual-care-and-telesurgery},
  note         = {Accessed: 2025-09-12}  
}

@article{mcbride2021detailed,
  title={Detailed cost of robotic-assisted surgery in the Australian public health sector: from implementation to a multi-specialty caseload},
  author={McBride, Kate and Steffens, Daniel and Stanislaus, Christina and Solomon, Michael and Anderson, Teresa and Thanigasalam, Ruban and Leslie, Scott and Bannon, Paul G},
  journal={BMC Health Services Research},
  volume={21},
  number={1},
  pages={108},
  year={2021},
  publisher={Springer}
}

@article{burke2024robotic,
  title={Robotic surgery in low-and middle-income countries},
  author={Burke, J and Gnanaraj, J and Dhanda, J and Martins, B and Vinck, EE and Saklani, A and Harji, D},
  journal={The Bulletin of the Royal College of Surgeons of England},
  volume={106},
  number={3},
  pages={138--141},
  year={2024},
  publisher={Royal College of Surgeons}
}

@article{neri2024novel,
  title={A novel affordable user interface for robotic surgery training: design, development and usability study},
  author={Neri, Alberto and Coduri, Mara and Penza, Veronica and Santangelo, Andrea and Oliveri, Alessandra and Turco, Enrico and Pizzirani, Mattia and Trinceri, Elisa and Soriero, Domenico and Boero, Federico and others},
  journal={Frontiers in Digital Health},
  volume={6},
  pages={1428534},
  year={2024},
  publisher={Frontiers Media SA}
}

@article{kanji2021room,
  title={Room size influences flow in robotic-assisted surgery},
  author={Kanji, Falisha and Cohen, Tara and Alfred, Myrtede and Caron, Ashley and Lawton, Samuel and Savage, Stephen and Shouhed, Daniel and Anger, Jennifer T and Catchpole, Ken},
  journal={International Journal of Environmental Research and Public Health},
  volume={18},
  number={15},
  pages={7984},
  year={2021},
  publisher={MDPI}
}

@incollection{zhao2023robotic,
  title={Robotic Operating Room Configuration},
  author={Zhao, Hang Yan and Zhan, Chunyan},
  booktitle={Pediatric Robotic Surgery},
  pages={17--20},
  year={2023},
  publisher={Springer}
}

@article{liu2024robotic,
  title={Robotic surgery in gastrointestinal surgery: history, current status, and challenges},
  author={Liu, Yu and Feng, Qingyang and Xu, Jianmin},
  journal={Intelligent Surgery},
  year={2024},
  publisher={Elsevier}
}

@article{clanahan2023does,
  title={How does robotic-assisted surgery change OR safety culture?},
  author={Clanahan, Julie M and Awad, Michael M},
  journal={AMA Journal of Ethics},
  volume={25},
  number={8},
  pages={615--623},
  year={2023},
  publisher={American Medical Association}
}

@incollection{hart1988development,
  title={Development of NASA-TLX (Task Load Index): Results of empirical and theoretical research},
  author={Hart, Sandra G and Staveland, Lowell E},
  booktitle={Advances in psychology},
  volume={52},
  pages={139--183},
  year={1988},
  publisher={Elsevier}
}

@inproceedings{saha2025based,
  title={Based: Bundle-adjusting surgical endoscopic dynamic video reconstruction using neural radiance fields},
  author={Saha, Shreya and Liang, Zekai and Lin, Shan and Lu, Jingpei and Yip, Michael and Liu, Sainan},
  booktitle={2025 IEEE/CVF Winter Conference on Applications of Computer Vision (WACV)},
  pages={3003--3012},
  year={2025},
  organization={IEEE}
}

@inproceedings{liang2025differentiable,
  title={Differentiable rendering-based pose estimation for surgical robotic instruments},
  author={Liang, Zekai and Chiu, Zih-Yun and Richter, Florian and Yip, Michael C},
  booktitle={2025 IEEE/RSJ International Conference on Intelligent Robots and Systems (IROS)},
  pages={20898--20905},
  year={2025},
  organization={IEEE}
}

@misc{cornerstone_robotics,
  author       = {{Cornerstone Robotics Limited}},
  title        = {Cornerstone Robotics},
  howpublished = {\url{https://en.csrbtx.com/}},
  note         = {Accessed: 2026-07-14}
}

\end{document}